\pdfoutput=1
\documentclass[11pt]{article}

\usepackage[]{acl}
\usepackage{times}
\usepackage{latexsym}

\usepackage[T1]{fontenc}
\usepackage[utf8]{inputenc}

\usepackage{microtype}

\usepackage{inconsolata}

\usepackage{adjustbox}
\usepackage{multirow}
\usepackage{booktabs}
\usepackage{amsmath}
\usepackage[section]{placeins}
\usepackage{hyperref}
\title{Practical Token Pruning for Foundation Models in Few-shot Conversational Virtual Assistant Systems}

  \author{\small
    Haode Qi\thanks{\, Equal contributions from the corresponding authors: \texttt{\{haode.qi,cheng.qian\}@ibm.com}.} , Cheng Qian$^{*}$, Jian Ni, Pratyush Singh, Reza Fazeli, Gengyu Wang, Zhongzheng Shu, Eric Wayne, Juergen Bross  \\
    \\
          IBM
}

\begin{document}
\maketitle
\begin{abstract}
In an enterprise Virtual Assistant (VA) system, intent classification is the crucial component that determines how a user input is handled based on what the user wants. The VA system is expected to be a cost-efficient SaaS service with low training and inference time while achieving high accuracy even with a small number of training samples. We pretrain a transformer-based sentence embedding model with a contrastive learning objective and leverage the model's embeddings as features when training intent classification models. Our approach achieves the state-of-the-art results for few-shot scenarios and performs better than other commercial solutions on popular intent classification benchmarks. However, generating features via a transformer-based model increases the inference time, especially for longer user inputs, due to the quadratic runtime of the transformer's attention mechanism. On top of model distillation, we introduce a practical multi-task adaptation approach that configures dynamic token pruning without the need for task-specific training for intent classification. We demonstrate that this approach improves the inference speed of popular sentence transformer models without affecting model performance.

% In an enterprise virtual assistant (VA) system, intent detection is the key component that allows identification of user intent given an input text. With the advancement of foundation models, enterprise customers expect an VA system to still train in 10-15 minutes for iterative editing, adaptable to any proprietary domain, support highly imbalance or extreme few shot training, and inference time in milliseconds. There are 4 primary challenges. 1. extreme few shot where customers only have a few examples to launch a new intent. This is typical when a new customer joins. On the other hand, customers can have very large amount of intents contrary to popular academic settings where a handful of intents are sampled for evaluation 3. fast multi-task adaptation that permits the model to adapt to new domains in minutes 4. low inference time for short and long utterances.

% Our contributions are the following:
% 1. all intent extreme few shot benchmark (with challenging test set)
% 2. practical multi-task token pruning for transformer model for production

% Notes:
% 1. we want to compare on challenging test set
% 2. explore what LLM does
% 3. start measuring train time & runtime

% our previous paper on similar topics:
% https://aclanthology.org/2021.naacl-industry.38.pdf

\end{abstract}

\section{Introduction}
Intent classification, the task of identifying the purpose of a user input utterance, is a crucial component in modern task-oriented virtual assistant(VA) systems. The detected intent, combined with extracted entities, is used to determine the proper dialog nodes in a pre-designed dialog tree to trigger corresponding responses to the end user \citep{wang-etal-2022-benchmarking}. The training time and single query inference time of the underlying intent classification algorithm both play an important role for the efficiency and end user experience of a VA system \citep{qi2020benchmarking}.
Additionally, although a large, balanced, and clean training set is desired for training machine learning algorithms in general, commercial VA systems are expected to handle training sets with limited size. Since good training data are costly to create, chatbot designers often start with only a few examples per intent during early phases of the bot design lifecycle or when new intents are added. This poses a challenge for the few-shot capabilities of intent classification algorithms deployed in commercial VA systems.  

The transformer architecture \citep{vaswani2017attention} has shown great performance in a wide range of tasks including intent classification. Supervised SimCSE \citep{gao2021simcse} shows that using positive sentences pairs from NLI \citep{gao2021simcse} substantially improves the performance of sentence embedding model with transformer architecture. We use a similar approach by gathering intents from different datasets across domains and treat examples from the same intent as positive pairs. This simple approach demonstrates strong performance in extreme few-shot intent classification scenario. However, the high accuracy comes at a cost of substantial compute resources and memory capacity, which can lead to a high latency or high operating expenses for commercial VA systems. In this paper, we propose a practical multitask token pruning procedure to significantly reduce the computational cost for transformer architectures, which does not require additional adaptation. We show the approach can work seamlessly with out-of-the-box sentence transformer models without affecting accuracy.
% These instructions are for authors submitting papers to *ACL conferences using \LaTeX. They are not self-contained. All authors must follow the general instructions for *ACL proceedings,\footnote{\url{http://acl-org.github.io/ACLPUB/formatting.html}} and this document contains additional instructions for the \LaTeX{} style files.

% The templates include the \LaTeX{} source of this document (\texttt{acl.tex}),
% the \LaTeX{} style file used to format it (\texttt{acl.sty}),
% an ACL bibliography style (\texttt{acl\_natbib.bst}),
% an example bibliography (\texttt{../custom.bib}),
% and the bibliography for the ACL Anthology (\texttt{anthology.bib}).

\section{Related Work}

% \subsection{Few shot intent classification}

% \subsection{Transformers and token pruning}
The transformer architecture \citep{vaswani2017attention} has become ubiquitous in various domains and applications due to its state-of-the-art performance. Different types of transformer-based models have been applied to intent classification in the context of VA systems. IntentBert \citep{zhang2021effectiveness} uses the representation of the [CLS] token from an encoder only BERT-based model \citep{devlin2019bert} to train a classifier head. Sentence BERT models \citep{DBLP:journals/corr/abs-1908-10084} which are based on a Siamese and triplet architecture are used as feature extractors for intent classification. Recently, \citet{parikh2023exploring} perform zero-shot intent classification by including a description of each intent class in the prompt for a large language model.
The state-of-the-art performance of these large transformer models comes with a challenge in terms of complexity, memory consumption, and scalability in real world production systems. 

Many approaches have been proposed to improve the efficiency of transformer-based models such as token pruning (\citealt{goyal2020powerbert}, \citealt{kim2021lengthadaptive}, \citealt{DBLP:journals/corr/abs-2105-11618}, \citealt{kim2022learned}, \citealt{wang2021spatten}), quantization (\citealt{DBLP:journals/corr/abs-1712-05877}, \citealt{lin2023awq}, \citealt{xiao2023smoothquant}, \citealt{dettmers2022llmint8}), distillation (\citealt{hinton2015distilling}, \citealt{mobahi2020selfdistillation}, \citealt{allenzhu2023understanding}, \citealt{sanh2020distilbert}), sparse and low-rank approximation (\citealt{li2023losparse}, \citealt{tahaei2021kroneckerbert}, \citealt{chen2021scatterbrain}, \citealt{wang2020linformer}), and more efficient implementation of the exact attention algorithm (\citealt{dao2022flashattention}). 

Token pruning aims to discard a subset of tokens which effectively decreases the input sequence length to achieve speedup. There are two main types of token pruning techniques. The first one introduces an additional procedure during pre-training or finetuning for adapting token pruning to a specific task or dataset (token-pruning-awared training). For example, \citet{goyal2020powerbert} introduce soft-extract layers with trainable-parameters into transformer encoder blocks during finetuning for a model to learn which tokens to drop. \citet{kim2021lengthadaptive} introduce LengthDrop and LayerDrop to the transformer architectures, which cannot be directly applied to an off-the-shelf pretrained transformer-based model without additional training or finetuning. Moreover, both of the aforementioned techniques prune all input sentences to the same length. In our production environment, the length of user utterances can vary from a single token to 1000+ characters, making these techniques unpractical for production intent classification systems. To overcome this, \citet{kim2022learned} propose a trainable threshold per layer to perform token pruning. These parameters are trained via adding a regularization term in addition to the original loss function. However, enterprise VA systems require minimal adaption for new tasks. This technique requires a new threshold to be learned for each new task, rendering it unsuitable for enterprise VA systems where the typical training time is less than 10 minutes. 

The second type of token pruning techniques can be applied directly to a trained model (post training token pruning). \citet{wang2021spatten} propose cascade token pruning. For each layer, there is a configurable ratio, based on which tokens will be pruned on the fly without additional training. However, configuring a parameter for each layer can be difficult for deep models over a wide variety of tasks. Additionally, while the technique reduces computation and memory consumption, they also point out that token pruning could increase memory access.

% Additionally, it also points out the additional memory access of token pruning that could limit the inference speed-up of the technique in practice.

\section{Sentence Embedding Model Pretraining and Optimization for Intent Classification}

In this section, we briefly describe the contrastive-learning training and optimization procedure and give an overview of the practical token pruning technique that has been deployed on our VA system in production. Even though this technique is applied to intent classification, it is task-agnostic and can be applied to any transformer-based model.
\subsection{Pretraining and Distillation}
We pretrain our sentence embedding model with Multiple Negative Loss \citep{journals/corr/HendersonASSLGK17} using proprietary intent classification datasets. We create positive example pairs $(a_i, p_i)$ by sampling from the same intent. We construct a collection of example pairs from intents across datasets from different domains. After pretraining, a student model with fewer layers is distilled from the teacher model using MSE Loss \citep{reimers2020making}. The token pruning procedure is applied to the student model deployed in production. We introduce this token pruning procedure for transformer models in the following subsections.
 %Let $I_i=\{(a_1, p_1), (a_2, p_2),  \ldots, (a_n, p_n)\}$

\subsection{Attention Mechanism}
The notation mostly follows the transformer paper \citep{vaswani2017attention}. Transformer layers are the building blocks of transformer-based architectures. In the context of NLP, a transformer layer takes a batch of tokenized and featurized sentences $X \in \mathcal{R}^{B \times n \times d_{model}}$ as input and outputs a tensor of the same size, where \(B\) is the batch size, \(n\) is the sequence length, and \(d_{model}\) is the embedding dimension.

Within each transformer layer, there are two sub-layers, a multi-head self-attention block, and a simple token-wise feed-forward network with two fully connected layers. Let \(H\) be the number of heads in the multi-head self-attention block. Each head is parameterized by three matrices $W_i^Q \in \mathcal{R}^{d_{model} \times d_k}$, $W_i^K \in \mathcal{R}^{d_{model} \times d_k}$, and $W_i^V \in \mathcal{R}^{d_{model} \times d_k}$, where \(i\) is the index for a head, and $d_k \times H = d_{model}$. An input to a self-attention head is linearly transformed into three matrices query \(Q\), key \(K\), and value \(V\) by the three parameter matrices, respectively. Attention scores and the final representation are calculated as 

\vspace{-0.1in}
\begin{equation}
\footnotesize{
A(Q, K) = softmax(\dfrac{QK^T}{\sqrt{d_k}})
}
\end{equation}
\vspace{-0.1in}

\vspace{-0.1in}
\begin{equation}
\footnotesize{
Attention(Q, K, V) = A(Q, K)V
}
\end{equation}
\vspace{-0.1in}

The complexity of a self-attention block is $O(n^2 \cdot d + d^2 \cdot n)$, where $d$ is the representation dimension.

\subsection{Proposed Method}

Token pruning aims to reduce the complexity of a self-attention block by discarding a subset of tokens, resulting in a quadratic reduction of the complexity of self-attention (and therefore all subsequent layers). The proposed method is configured with three parameters:  \(\it{s}\) denotes the minimum length of a sequence (measured by the number of tokens) to which pruning will be applied; \(\it{q}\) denotes the quantile percent which defines a threshold on token importance to discard tokens, and \(\it{l}\) denotes the index of the layer where token pruning is applied.

\subsubsection{Attention score based token importance}
Following existing works, we measure the importance of a token based on attention scores. Formally, given an attention head $h$, the attention score $A_{i, j}^h$ measures the amount of attention a token $i$ has on token $j$. At a transformer layer, we first average all $H$ attention score matrices from the $H$ attention heads. The averaged attention score matrix is denoted as $\overline{A}$. Secondly, for each token $j$, the final importance measurement is calculated as 

\vspace{-0.1in}
\begin{equation}
\footnotesize{
\sum_i \overline{A}_{i, j}
}
\end{equation}
\vspace{0.0in}
where the summation is taken over all the non-padded tokens. It's worth noting that the computation of attention score based token importance should not involve any padded tokens, so that the importance scores for tokens in the same sentence are consistent regardless of the max length of all sentences within the same batch. With this definition of token importance, for each sentence, only the tokens with the highest top \(\it{q}\) importance scores will be kept. For a tokenized sentence of length \(n\), whose matrix representation is of shape ${1 \times n \times d}$, the corresponding output from a token-pruned self-attention block would be of shape ${1 \times n^{\prime} \times d}$, where $n^{\prime} \leq n$ and 1 being the placeholder for batch size dimension.

\subsubsection{Protection against excessive pruning}

Through experiments, we notice performing token pruning at a single early transformer layer usually suffices to achieve a good trade-off between inference speed-up and accuracy. By pruning at an early layer $l$, all subsequent layers will benefit in terms of inference speed-up, because during forward-pass, the pruned tokens will not reach the subsequent layers after the layer where they have been pruned. By limiting the number of layers we apply token pruning to, we also limit the additional overhead from performing token pruning: calculation and sorting of averaged attention scores on non-padded tokens, and removal of the pruned tokens. 
% For all experiments in this work, we only prune the first transformer layer.

Based on traffic patterns in our VA system, it is common to see short sentences with less than ten tokens. For such short sentences, the quadratic complexity of transformer blocks is less of a concern. In addition, to prevent important information from being discarded during token pruning, we aim to configure $s$, the minimum number of tokens in a sentence for it to be token pruned, to balance between speedup and preserving information.

\subsubsection{Practical multitask adaptation}
\label{section:setting}
Enterprise VA systems are expected to serve a large number of customers from a wide variety of domains. Therefore, any algorithms applied are expected to generalize well on datasets of different characteristics, such as number of training examples, average length of training examples and the domains of these examples. When applying dynamic token pruning technique to enterprise VA systems, it is desirable to devise with a configuration that works for a wide variety of tasks across different domains out of the box. 

We propose a simple and robust post-training adaptation approach that requires no further tuning of the token pruning configuration for new tasks. We perform hyperparameter search on a hold-out set of intent classification datasets for the optimal token pruning configuration for our production VA system. To illustrate this process, we introduce some additional notations: a transformer model $E$ is used as a feature extractor, on top of which we train a classifier head $f$ for each dataset, and evaluation metric for intent classification such as accuracy and F1 is denoted as $M$. For a hold-out collection of $n$ intent classification tasks $T=\{(t_1, d_1), (t_2, d_2),  \ldots, (t_n, d_n)\}$ where $(t_i, d_i)$ is the training and development set pair from $task_i$, we evaluate combinations of $(s, q, l)$  iteratively on all tasks and find the one that optimizes

\vspace{-0.1in}
\begin{equation}
s^*, q^*, l^* =  \text{argmax}_{s, q, l} \frac{1}{n} \sum_{i=1}^{n} M(f(E(t_i | s, q, l)), d_i).  
\end{equation}
\vspace{-0.1in}

% For our production model, we fit a single classification layer on top of the embedding produced by the sentence embedding model for each hold-out dataset and use classification accuracy as the evaluation metric.
% The post-training adaptation is conducted in an offline fashion on a hold-out set. 

The resulting set of optimal configuration $s^*, q^*, l^*$ will be applied to all unseen tasks. 

\section{Experiments}

In this section, we conduct three sets of experiments. We first compare our production system  \footnote{Accessible through \href{https://www.ibm.com/products/watsonx-assistant}{UI} and \href{https://cloud.ibm.com/apidocs/assistant-v2}{API}. In this paper, we refer to two versions of intent classification algorithms available in our product: 15-Apr-2023 (\textbf{Latest}) and 20-Dec-2022 (\textbf{Previous}). \textbf{Latest} uses transformer-based models with the token pruning described in this paper, thus being referred to as \textbf{IBM Watsonx(LM)} in the following sections. For more detail about model versioning, refer to \href{https://cloud.ibm.com/docs/assistant?topic=assistant-algorithm-version}{our documentation.} }with the proposed optimizations against both academic and commercial intent classification algorithms. Then we extend beyond our product and apply our token pruning procedure to an open-sourced transformer model and verify its general efficacy and applicability.

\subsection{Experiment I: Comparing against academic intent classification algorithms}

We compare our deployed production intent classification algorithm against the following recent academic intent classification  baselines. \textbf{IsoIntentBert} \citep{zhang2022finetuning} further pretrains a BERT model \citep{devlin2019bert} on part of the CLINC150 \citep{larson-etal-2019-evaluation} dataset with additional regularization terms. \textbf{SBERT-NLI}, \textbf{SBERT-para}, and \textbf{SimCSE-NLI} \citep{ma-etal-2022-intentemb} utilize sentence encoders for intent classification. In our experiment, these aforementioned models are used as feature extractors in inference mode then we train linear classifier heads with the extracted features. \textbf{DFT++} \citep{zhang2023revisit} directly finetunes a model with augmented unlabeled dataset (\textbf{C}ontext \textbf{A}ugmentation) and then performs Sequential Self-Distillation (\textbf{SSD}). For \textbf{DFT++}, we finetune a BERT model for each dataset. Due to the code of \textbf{DFT++} not being released yet as of the date of writing this paper, we are only able to implement a working version of \textbf{DFT++ (w/SSD)} (but not the full \textbf{DFT++  (w/CA, SSD)}), following the original paper on a best effort basis. For other models, we download the released academic code for our experiments.

\subsubsection{Datasets and metric}
% add hint3??
Experiments are conducted with three intent classification datasets: BANKING77 \citep{zhang2022pretrained}, CLINC150 \citep{larson-etal-2019-evaluation}, and HWU64 \citep{liu2019benchmarking}. Following \citet{wang-etal-2022-benchmarking}, we randomly sample 1-shot, 2-shot, 3-shot, and 5-shot variants of the training sets; and randomly sample more difficult versions of test sets based on jaccard distance and tf-idf. We use accuracy as the metric.

\subsubsection{Results}

Out of the total of 36 settings reported in Table \ref{tab:main}, our system performs the best in 24 of them. 

\begin{table*}[]
% \begin{table*}[!htb]
\begin{adjustbox}{max width=0.85\textwidth}
\begin{tabular}{c|l|l|ccc}
\toprule
\multirow{2}{*}{\textbf{Training set variant}} & \multirow{2}{*}{\textbf{Dataset}} & \multirow{2}{*}{\textbf{Algorithm}} & \multicolumn{3}{c}{Accuracy}  \\
                              &                  &                      & \textbf{Full test set} & \textbf{Difficult test set(jaccard)} & \textbf{Difficult test set(tfidf)} \\
\midrule
\multirow{17}{*}{1-shot}                        & \multirow{6}{*}{BANKING77}        & DFT++(w/SSD)         & 28.60\%                & 25.22\%                              & 18.65\%                            \\
                              &                  & IsoIntentBert        & 40.97\%                & 35.97\%                              & 32.21\%                            \\
                              &                  & SBERT-NLI            & 43.21\%                & 37.19\%                              & 35.32\%                            \\
                              &                  & SBERT-para           & 46.55\%                & 40.57\%                              & \textbf{36.49}\%                            \\
                              &                  & SimCSE-NLI           & 42.62\%                & 36.28\%                              & 34.55\%                            \\
                              &                  & IBM Watsonx(LM)                 & \textbf{47.72}\%                & \textbf{42.73}\%                              & 34.68\%                            \\ \cmidrule{2-6}
                              & \multirow{5}{*}{CLINC150}         & DFT++(w/SSD)         & 48.87\%                & 43.81\%                              & 34.43\%                            \\
                              &                  & SBERT-NLI            & 58.71\%                & 55.20\%                              & 45.07\%                            \\
                              &                  & SBERT-para           & 59.66\%                & 52.13\%                              & 47.73\%                            \\
                              &                  & SimCSE-NLI           & 57.75\%                & 51.47\%                              & 45.73\%                            \\
                              &                  & IBM Watsonx(LM)                 & \textbf{67.40}\%                & \textbf{60.80}\%                              & \textbf{54.00}\%                            \\ \cmidrule{2-6}
                              & \multirow{6}{*}{HWU64}            & DFT++(w/SSD)         & 40.48\%                & 40.55\%                              & 36.39\%                            \\
                              &                  & IsoIntentBert        & \textbf{56.34}\%                & \textbf{54.87}\%                              & \textbf{50.81}\%                            \\
                              &                  & SBERT-NLI            & 46.56\%                & 44.35\%                              & 41.13\%                            \\
                              &                  & SBERT-para           & 44.14\%                & 44.35\%                              & 40.97\%                            \\
                              &                  & SimCSE-NLI           & 44.79\%                & 43.71\%                              & 40.81\%                            \\
                              &                  & IBM Watsonx(LM)                 & 52.78\%                & 50.00\%                              & 47.74\%                            \\ \midrule
\multirow{17}{*}{2-shot}                        & \multirow{6}{*}{BANKING77}        & DFT++(w/SSD)         & 53.05\%                & 43.25\%                              & 36.00\%                            \\
                              &                  & IsoIntentBert        & 55.20\%                & 46.39\%                              & 39.53\%                            \\
                              &                  & SBERT-NLI            & 60.42\%                & 49.80\%                              & 44.68\%                            \\
                              &                  & SBERT-para           & 62.33\%                & \textbf{53.45}\%                              & 46.62\%                            \\
                              &                  & SimCSE-NLI           & 59.18\%                & 51.24\%                              & 42.99\%                            \\
                              &                  & IBM Watsonx(LM)                 & \textbf{63.02}\%                & 51.95\%                              & \textbf{46.88}\%                            \\ \cmidrule{2-6}
                              & \multirow{5}{*}{CLINC150}         & DFT++(w/SSD)         & 71.92\%                & 61.84\%                              & 52.24\%                            \\
                              &                  & SBERT-NLI            & 71.75\%                & 65.60\%                              & 56.00\%                            \\
                              &                  & SBERT-para           & 73.17\%                & \textbf{67.07}\%                              & 59.73\%                            \\
                              &                  & SimCSE-NLI           & 71.26\%                & 65.87\%                              & 57.87\%                            \\
                              &                  & IBM Watsonx(LM)                 & \textbf{78.13}\%                & 66.27\%                              & \textbf{61.60}\%                            \\ \cmidrule{2-6}
                              & \multirow{6}{*}{HWU64}            & DFT++(w/SSD)         & 62.29\%                & 58.87\%                              & 53.58\%                            \\
                              &                  & IsoIntentBert        & 66.78\%                & 65.26\%                              & 59.26\%                            \\
                              &                  & SBERT-NLI            & 64.49\%                & 60.65\%                              & 55.65\%                            \\
                              &                  & SBERT-para           & 63.29\%                & 58.55\%                              & 53.39\%                            \\
                              &                  & SimCSE-NLI           & 63.66\%                & 60.32\%                              & 56.13\%                            \\
                              &                  & IBM Watsonx(LM)                 & \textbf{68.68}\%                & \textbf{66.29}\%                              & \textbf{61.13}\%                            \\ \midrule
\multirow{17}{*}{3-shot}                        & \multirow{6}{*}{BANKING77}        & DFT++(w/SSD)         & 59.82\%                & 49.45\%                              & 45.06\%                            \\
                              &                  & IsoIntentBert        & 59.35\%                & 50.18\%                              & 45.19\%                            \\
                              &                  & SBERT-NLI            & 61.81\%                & 50.33\%                              & 45.58\%                            \\
                              &                  & SBERT-para           & 64.18\%                & 53.06\%                              & 46.49\%                            \\
                              &                  & SimCSE-NLI           & 59.44\%                & 48.89\%                              & 43.77\%                            \\
                              &                  & IBM Watsonx(LM)                 & \textbf{65.29}\%                & \textbf{54.29}\%                              & \textbf{50.00}\%                            \\ \cmidrule{2-6}
                              & \multirow{5}{*}{CLINC150}         & DFT++(w/SSD)         & 78.70\%                & 70.21\%                              & 61.12\%                            \\
                              &                  & SBERT-NLI            & 77.42\%                & 70.80\%                              & 64.67\%                            \\
                              &                  & SBERT-para           & 78.24\%                & 72.93\%                              & 67.60\%                            \\
                              &                  & SimCSE-NLI           & 77.39\%                & 70.27\%                              & 64.80\%                            \\
                              &                  & IBM Watsonx(LM)                 & \textbf{82.91}\%                & \textbf{74.93}\%                              & \textbf{68.53}\%                            \\ \cmidrule{2-6}
                              & \multirow{6}{*}{HWU64}            & DFT++(w/SSD)         & 66.97\%                & 67.00\%                              & 59.39\%                            \\
                              &                  & IsoIntentBert        & \textbf{71.00}\%                & \textbf{70.45}\%                              & \textbf{63.74}\%                            \\
                              &                  & SBERT-NLI            & 68.68\%                & 68.23\%                              & 61.77\%                            \\
                              &                  & SBERT-para           & 69.24\%                & 67.74\%                              & 60.81\%                            \\
                              &                  & SimCSE-NLI           & 66.72\%                & 65.16\%                              & 60.32\%                            \\
                              &                  & IBM Watsonx(LM)                 & 70.50\%                & 69.35\%                              & 63.71\%                            \\ \midrule
\multirow{17}{*}{5-shot}                        & \multirow{6}{*}{BANKING77}        & DFT++(w/SSD)         & 75.08\%                & 64.23\%                              & 55.53\%                            \\
                              &                  & IsoIntentBert        & 69.19\%                & 59.12\%                              & 51.56\%                            \\
                              &                  & SBERT-NLI            & 72.92\%                & 62.16\%                              & 56.36\%                            \\
                              &                  & SBERT-para           & 73.47\%                & 63.59\%                              & 54.29\%                            \\
                              &                  & SimCSE-NLI           & 70.75\%                & 60.99\%                              & 54.03\%                            \\
                              &                  & IBM Watsonx(LM)                 & \textbf{76.88}\%                & \textbf{67.66}\%                              & \textbf{59.35}\%                            \\ \cmidrule{2-6}
                              & \multirow{5}{*}{CLINC150}         & DFT++(w/SSD)         & 85.16\%                & 76.16\%                              & 67.60\%                            \\
                              &                  & SBERT-NLI            & 82.02\%                & 76.67\%                              & 68.93\%                            \\
                              &                  & SBERT-para           & 83.44\%                & 76.93\%                              & 70.27\%                            \\
                              &                  & SimCSE-NLI           & 81.28\%                & 76.80\%                              & 68.67\%                            \\
                              &                  & IBM Watsonx(LM)                 & \textbf{88.53}\%                & \textbf{80.93}\%                              & \textbf{71.73}\%                            \\ \cmidrule{2-6}
                              & \multirow{6}{*}{HWU64}            & DFT++(w/SSD)         & \textbf{77.62}\%                & \textbf{76.00}\%                              & \textbf{71.68}\%                            \\
                              &                  & IsoIntentBert        & 74.52\%                & 72.29\%                              & 67.29\%                            \\
                              &                  & SBERT-NLI            & 73.69\%                & 74.52\%                              & 66.77\%                            \\
                              &                  & SBERT-para           & 74.44\%                & 74.52\%                              & 67.58\%                            \\
                              &                  & SimCSE-NLI           & 74.35\%                & 74.35\%                              & 67.58\%                            \\
                              &                  & IBM Watsonx(LM)                 & 76.39\%                & 75.32\%                              & 70.16\%  \\            
                              \bottomrule
\end{tabular}

\end{adjustbox}
\caption{Accuracy comparison between our solution (Ours) vs academic intent classifiers on few shot scenario. Best results are in bold.}
\label{tab:main}
\end{table*}

\begin{table*}[]
    \centering
    \begin{adjustbox}{max width=1\textwidth}
    \begin{tabular}{|l|l|l|c|c|c|c|}
    \toprule
        DATASET & Training set variant & Token Pruned & Accuracy(Logistic regression head) &  Time(to generate embeddings for training set)  & Change in accuracy & Speed up  \\ 
    \midrule
        \multirow{4}{*}{CLINC150} & 3-shot & N & 79.33\% &  1.30  & ~ &   \\ 
        ~ & 5-shot & N & 84.73\% &  3.02  & ~ &   \\ 
        ~ & 3-shot & Y & 80.91\% &  1.01  & 1.58\% & 23\%  \\ 
        ~ & 5-shot & Y & 87.47\% &  2.00  & 2.73\% & 34\%  \\ 
    \midrule
        \multirow{4}{*}{HWU64} & 3-shot & N & 67.47\% &  0.49  & ~ &   \\ 
        ~ & 5-shot & N & 76.77\% &  0.84  & ~ &   \\
        ~ & 3-shot & Y & 70.26\% &  0.39  & 2.79\% & 20\%  \\ 
        ~ & 5-shot & Y & 76.30\% &  0.64  & -0.46\% & 24\%  \\ 
    \midrule
        \multirow{4}{*}{BANKING77} & 3-shot & N & 68.57\% &  1.81  & ~ &   \\ 
        ~ & 5-shot & N & 77.47\% &  4.00  & ~ &   \\ 
        ~ & 3-shot & Y & 71.62\% &  1.24  & 3.05\% & 31\%  \\ 
        ~ & 5-shot & Y & 79.35\% &  2.67  & 1.88\% & 33\% \\ 
    \bottomrule
    \end{tabular}

    \end{adjustbox}
    \caption{Comparing MiniLM-L12 vs token pruned MiniLM-L12 on intent classification accuracy and time (in seconds) taken to generate sentence embeddings for training set.  }
    \label{tab:minilm}
\end{table*}

\subsection{Experiment II: Comparing against commercial solutions}

We compare our system with other commercial solutions on intent classification performance. 
% Intent classification is an important part of commercial VA solutions, as it's a common practice for chatbot designers to define chatbot responses or actions based on the predicted intent, along with the extracted slot information. Few-shot intent classification performance is especially important setting for commercial VA solutions since creating training set is time consuming and an easy on-boarding is crucial for customer experience. 

\subsubsection{Dataset and metrics}
We conduct benchmarking on the 3 commonly used intent classification datasets HWU64, BANKING77, and CLINC150. For each dataset, we use the 10-shot version of the training set following \citet{MehriDialoGLUE2020}. For evaluation, we measure weighted averages of precision, recall, and F1 following the setup of a recent Cognigy blog\footnote{https://www.cognigy.com/blog/benchmarking-nlu-engines-comparing-market-leaders \label{fnlabel}}.

\subsubsection{Results and Analysis}

We add the metrics from our system to the comparison in the \hyperref[fnlabel]{Cognigy blog} and put the best (highest) metrics in bold in Table \ref{tab:commercial}. Our system is consistently the best across the 3 datasets and metrics.

\begin{table}[h]
    \centering
    \begin{adjustbox}{max width=0.5\textwidth}

    \begin{tabular}{|l|l|lll|}
    \toprule
        DATASET  & Solution Provider  & precision & recall & F1  \\ 
          \midrule
        \multirow{5}{*}{HWU64}   
          & Cognigy  & \bf{0.84} & 0.79 & 0.8  \\ 
          & Microsoft CLU  & 0.81 & 0.8 & 0.79  \\ 
          & Google Dialogflow  & 0.74 & 0.63 & 0.66  \\ 
          & IBM Watsonx(Previous)  & \bf{0.84} & 0.75 & 0.78  \\ 
          & IBM Watsonx(LM)  & \bf{0.84} & \bf{0.82} & \bf{0.82}  \\ 
          \midrule
        \multirow{5}{*}{BANKING77}  
          & Cognigy  & 0.81 & 0.8 & 0.8  \\ 
          & Microsoft CLU  & 0.74 & 0.7 & 0.7  \\ 
          & Google Dialogflow  & 0.76 & 0.74 & 0.74  \\ 
          & IBM Watsonx(Previous) & 0.82 & 0.81 & 0.81  \\ 
          & IBM Watsonx(LM)  & \bf{0.86} & \bf{0.85} & \bf{0.85}  \\ 
          \midrule
        \multirow{5}{*}{CLINC150}  
          & Cognigy  & 0.91 & 0.87 & 0.88  \\ 
          & Microsoft CLU  & 0.86 & 0.84 & 0.84  \\
          & Google Dialogflow  & 0.8 & 0.76 & 0.77  \\ 
          & IBM Watsonx(Previous)  & 0.89 & 0.87 & 0.87  \\           
          & IBM Watsonx(LM) & \bf{0.92} & \bf{0.91} & \bf{0.91} \\ 
    \bottomrule
    \end{tabular}
    \end{adjustbox}
    \caption{Weighted average precision, recall, and F1 on 10-shot datasets by different commercial VA solutions. Note: Results for all methods except IBM Watsonx(LM) are obtained from the \hyperref[fnlabel]{Cognigy blog}.}
    \label{tab:commercial}
\end{table}

\subsection{Experiment III: Applying our token pruning technique to open-sourced transformer models for intent classification}

% In this section, we apply the same token pruning technique with configuration listed in Subsection \ref{section:setting} which is very similar to our production setting to an open-sourced transformer model, which is used as feature extractor for intent classification.

% \subsubsection{MiniLM-L12 for intent classification}
We apply our token pruning technique to a common sentence embedding model MiniLM-L12 (\citet{reimers-2019-sentence-bert}, \citet{wang2020minilm})\footnote{Weights are downloaded from https://huggingface.co/sentence-transformers/all-MiniLM-L12-v2} and study the trade-off between inference speed and intent classification accuracy. We use the off-the-shelf pre-trained MiniLM-L12 (with and without token pruning) as a feature extractor without finetuning, and use the extracted sentence embeddings to train logistic classifiers. The configuration for token pruning the MiniLM model is obtained by conducting the offline multitask adaptation described in Subsection \ref{section:setting} on an internally curated hold-out set of intent classification tasks using a smaller MiniLM model. We set the search space of $q$ to discrete values ranging from 0.6 to 0.8, $l$ from $\{1, 2, 3\}$ (only one of the early transformer layers will be pruned), and $s$ from 10 to 20 during post-training adaptation. The resulting optimal configuration $s, q, l$ is  $15, 0.8, 1$. We measure both accuracy and the time taken to produce sentence embeddings, including tokenizing, forward-passing, mean pooling, and normalizing. The time is averaged across 7 runs using 4 cores of Intel(R) Xeon(R) Gold 6258R CPU @ 2.70GHz.

% for each few-shot training set in one batch using 4 cores of Intel(R) Xeon(R) Gold 6258R CPU @ 2.70GHz. We repeat the time measurement for 7 times.

% \subsubsection{Token pruning configuration}  \label{minilm_config}
% It takes 3 parameters to configure the proposed token pruning technique. For the experiment with miniLM-L12, we set the \(\it{min\_sequence\_length}\) to 15, \(\it{quantile\_percent}\) to 0.8, and \(\it{prune\_layer\_num}\) to 1.

\subsubsection{Result and Analysis}

From the results in Table \ref{tab:minilm}, we notice that by applying our token pruning technique to MiniLM-L12, we achieve a speed-up from 20\% to 34\% without much loss in accuracy. In some settings, we even observe an increased accuracy with token pruning, which might be attributed to the removal of unimportant tokens. This experiment is an example that the proposed technique can be extended beyond our product and applied to open-sourced off-the-shelf transformer models for intent classification tasks. 

\section{Limitations}
The study is subject to a few limitations. (1) Our benchmark only concerns few-shot cases that do not fully reflect the challenges of a production VA system.  In production, the lack of training samples in some intents is also a symptom of data imbalance. (2) We only study the inference speedup using environments similar to our production. The inference speedup of the token pruning technique could vary depending on the implementation and hardware architecture.

% This document has been adapted
% by Steven Bethard, Ryan Cotterell and Rui Yan
% from the instructions for earlier ACL and NAACL proceedings, including those for 
% ACL 2019 by Douwe Kiela and Ivan Vuli\'{c},
% NAACL 2019 by Stephanie Lukin and Alla Roskovskaya, 
% ACL 2018 by Shay Cohen, Kevin Gimpel, and Wei Lu, 
% NAACL 2018 by Margaret Mitchell and Stephanie Lukin,
% Bib\TeX{} suggestions for (NA)ACL 2017/2018 from Jason Eisner,
% ACL 2017 by Dan Gildea and Min-Yen Kan, 
% NAACL 2017 by Margaret Mitchell, 
% ACL 2012 by Maggie Li and Michael White, 
% ACL 2010 by Jing-Shin Chang and Philipp Koehn, 
% ACL 2008 by Johanna D. Moore, Simone Teufel, James Allan, and Sadaoki Furui, 
% ACL 2005 by Hwee Tou Ng and Kemal Oflazer, 
% ACL 2002 by Eugene Charniak and Dekang Lin, 
% and earlier ACL and EACL formats written by several people, including
% John Chen, Henry S. Thompson and Donald Walker.
% Additional elements were taken from the formatting instructions of the \emph{International Joint Conference on Artificial Intelligence} and the \emph{Conference on Computer Vision and Pattern Recognition}.

% Entries for the entire Anthology, followed by custom entries
 \clearpage
\bibliography{anthology,custom}

\begin{thebibliography}{37}
\expandafter\ifx\csname natexlab\endcsname\relax\def\natexlab#1{#1}\fi

\bibitem[{Allen-Zhu and Li(2023)}]{allenzhu2023understanding}
Zeyuan Allen-Zhu and Yuanzhi Li. 2023.
\newblock \href {http://arxiv.org/abs/2012.09816} {Towards understanding
  ensemble, knowledge distillation and self-distillation in deep learning}.

\bibitem[{Chen et~al.(2021)Chen, Dao, Winsor, Song, Rudra, and
  Ré}]{chen2021scatterbrain}
Beidi Chen, Tri Dao, Eric Winsor, Zhao Song, Atri Rudra, and Christopher Ré.
  2021.
\newblock \href {http://arxiv.org/abs/2110.15343} {Scatterbrain: Unifying
  sparse and low-rank attention approximation}.

\bibitem[{Dao et~al.(2022)Dao, Fu, Ermon, Rudra, and
  Ré}]{dao2022flashattention}
Tri Dao, Daniel~Y. Fu, Stefano Ermon, Atri Rudra, and Christopher Ré. 2022.
\newblock \href {http://arxiv.org/abs/2205.14135} {Flashattention: Fast and
  memory-efficient exact attention with io-awareness}.

\bibitem[{Dettmers et~al.(2022)Dettmers, Lewis, Belkada, and
  Zettlemoyer}]{dettmers2022llmint8}
Tim Dettmers, Mike Lewis, Younes Belkada, and Luke Zettlemoyer. 2022.
\newblock \href {http://arxiv.org/abs/2208.07339} {Llm.int8(): 8-bit matrix
  multiplication for transformers at scale}.

\bibitem[{Devlin et~al.(2019)Devlin, Chang, Lee, and
  Toutanova}]{devlin2019bert}
Jacob Devlin, Ming-Wei Chang, Kenton Lee, and Kristina Toutanova. 2019.
\newblock \href {http://arxiv.org/abs/1810.04805} {Bert: Pre-training of deep
  bidirectional transformers for language understanding}.

\bibitem[{Gao et~al.(2021)Gao, Yao, and Chen}]{gao2021simcse}
Tianyu Gao, Xingcheng Yao, and Danqi Chen. 2021.
\newblock {SimCSE}: Simple contrastive learning of sentence embeddings.
\newblock In \emph{Empirical Methods in Natural Language Processing (EMNLP)}.

\bibitem[{Goyal et~al.(2020)Goyal, Choudhury, Raje, Chakaravarthy, Sabharwal,
  and Verma}]{goyal2020powerbert}
Saurabh Goyal, Anamitra~R. Choudhury, Saurabh~M. Raje, Venkatesan~T.
  Chakaravarthy, Yogish Sabharwal, and Ashish Verma. 2020.
\newblock \href {http://arxiv.org/abs/2001.08950} {Power-bert: Accelerating
  bert inference via progressive word-vector elimination}.

\bibitem[{Henderson et~al.(2017)Henderson, Al-Rfou, Strope, Sung, Lukács, Guo,
  Kumar, Miklos, and Kurzweil}]{journals/corr/HendersonASSLGK17}
Matthew~L. Henderson, Rami Al-Rfou, Brian Strope, Yun-Hsuan Sung, László
  Lukács, Ruiqi Guo, Sanjiv Kumar, Balint Miklos, and Ray Kurzweil. 2017.
\newblock \href
  {http://dblp.uni-trier.de/db/journals/corr/corr1705.html#HendersonASSLGK17}
  {Efficient natural language response suggestion for smart reply.}
\newblock \emph{CoRR}, abs/1705.00652.

\bibitem[{Hinton et~al.(2015)Hinton, Vinyals, and Dean}]{hinton2015distilling}
Geoffrey Hinton, Oriol Vinyals, and Jeff Dean. 2015.
\newblock \href {http://arxiv.org/abs/1503.02531} {Distilling the knowledge in
  a neural network}.

\bibitem[{Jacob et~al.(2017)Jacob, Kligys, Chen, Zhu, Tang, Howard, Adam, and
  Kalenichenko}]{DBLP:journals/corr/abs-1712-05877}
Benoit Jacob, Skirmantas Kligys, Bo~Chen, Menglong Zhu, Matthew Tang, Andrew~G.
  Howard, Hartwig Adam, and Dmitry Kalenichenko. 2017.
\newblock \href {http://arxiv.org/abs/1712.05877} {Quantization and training of
  neural networks for efficient integer-arithmetic-only inference}.
\newblock \emph{CoRR}, abs/1712.05877.

\bibitem[{Kim and Cho(2021)}]{kim2021lengthadaptive}
Gyuwan Kim and Kyunghyun Cho. 2021.
\newblock \href {http://arxiv.org/abs/2010.07003} {Length-adaptive transformer:
  Train once with length drop, use anytime with search}.

\bibitem[{Kim et~al.(2022)Kim, Shen, Thorsley, Gholami, Kwon, Hassoun, and
  Keutzer}]{kim2022learned}
Sehoon Kim, Sheng Shen, David Thorsley, Amir Gholami, Woosuk Kwon, Joseph
  Hassoun, and Kurt Keutzer. 2022.
\newblock \href {http://arxiv.org/abs/2107.00910} {Learned token pruning for
  transformers}.

\bibitem[{Larson et~al.(2019)Larson, Mahendran, Peper, Clarke, Lee, Hill,
  Kummerfeld, Leach, Laurenzano, Tang, and Mars}]{larson-etal-2019-evaluation}
Stefan Larson, Anish Mahendran, Joseph~J. Peper, Christopher Clarke, Andrew
  Lee, Parker Hill, Jonathan~K. Kummerfeld, Kevin Leach, Michael~A. Laurenzano,
  Lingjia Tang, and Jason Mars. 2019.
\newblock \href {https://doi.org/10.18653/v1/D19-1131} {An evaluation dataset
  for intent classification and out-of-scope prediction}.
\newblock In \emph{Proceedings of the 2019 Conference on Empirical Methods in
  Natural Language Processing and the 9th International Joint Conference on
  Natural Language Processing (EMNLP-IJCNLP)}, pages 1311--1316, Hong Kong,
  China. Association for Computational Linguistics.

\bibitem[{Li et~al.(2023)Li, Yu, Zhang, Liang, He, Chen, and
  Zhao}]{li2023losparse}
Yixiao Li, Yifan Yu, Qingru Zhang, Chen Liang, Pengcheng He, Weizhu Chen, and
  Tuo Zhao. 2023.
\newblock \href {http://arxiv.org/abs/2306.11222} {Losparse: Structured
  compression of large language models based on low-rank and sparse
  approximation}.

\bibitem[{Lin et~al.(2023)Lin, Tang, Tang, Yang, Dang, Gan, and
  Han}]{lin2023awq}
Ji~Lin, Jiaming Tang, Haotian Tang, Shang Yang, Xingyu Dang, Chuang Gan, and
  Song Han. 2023.
\newblock \href {http://arxiv.org/abs/2306.00978} {Awq: Activation-aware weight
  quantization for llm compression and acceleration}.

\bibitem[{Liu et~al.(2019)Liu, Eshghi, Swietojanski, and
  Rieser}]{liu2019benchmarking}
Xingkun Liu, Arash Eshghi, Pawel Swietojanski, and Verena Rieser. 2019.
\newblock \href {http://arxiv.org/abs/1903.05566} {Benchmarking natural
  language understanding services for building conversational agents}.

\bibitem[{Ma et~al.(2022)Ma, Wu, Yu, Zhao, and Lin}]{ma-etal-2022-intentemb}
Tingting Ma, Qianhui Wu, Zhiwei Yu, Tiejun Zhao, and Chin-Yew Lin. 2022.
\newblock \href {https://openreview.net/pdf?id=SzGx4ZQfHZq} {On the
  effectiveness of sentence encoding for intent detection meta-learning}.
\newblock In \emph{2022 Annual Conference of the North American Chapter of the
  Association for Computational Linguistics (NAACL 2022)}. Association for
  Computational Linguistics.

\bibitem[{Mehri et~al.(2020)Mehri, Eric, and Hakkani-Tur}]{MehriDialoGLUE2020}
S.~Mehri, M.~Eric, and D.~Hakkani-Tur. 2020.
\newblock Dialoglue: A natural language understanding benchmark for
  task-oriented dialogue.
\newblock \emph{ArXiv}, abs/2009.13570.

\bibitem[{Mobahi et~al.(2020)Mobahi, Farajtabar, and
  Bartlett}]{mobahi2020selfdistillation}
Hossein Mobahi, Mehrdad Farajtabar, and Peter~L. Bartlett. 2020.
\newblock \href {http://arxiv.org/abs/2002.05715} {Self-distillation amplifies
  regularization in hilbert space}.

\bibitem[{Parikh et~al.(2023)Parikh, Vohra, Tumbade, and
  Tiwari}]{parikh2023exploring}
Soham Parikh, Quaizar Vohra, Prashil Tumbade, and Mitul Tiwari. 2023.
\newblock \href {http://arxiv.org/abs/2305.07157} {Exploring zero and few-shot
  techniques for intent classification}.

\bibitem[{Qi et~al.(2020)Qi, Pan, Sood, Shah, Kunc, Yu, and
  Potdar}]{qi2020benchmarking}
Haode Qi, Lin Pan, Atin Sood, Abhishek Shah, Ladislav Kunc, Mo~Yu, and Saloni
  Potdar. 2020.
\newblock Benchmarking commercial intent detection services with
  practice-driven evaluations.
\newblock \emph{arXiv preprint arXiv:2012.03929}.

\bibitem[{Reimers and
  Gurevych(2019{\natexlab{a}})}]{DBLP:journals/corr/abs-1908-10084}
Nils Reimers and Iryna Gurevych. 2019{\natexlab{a}}.
\newblock \href {http://arxiv.org/abs/1908.10084} {Sentence-bert: Sentence
  embeddings using siamese bert-networks}.
\newblock \emph{CoRR}, abs/1908.10084.

\bibitem[{Reimers and
  Gurevych(2019{\natexlab{b}})}]{reimers-2019-sentence-bert}
Nils Reimers and Iryna Gurevych. 2019{\natexlab{b}}.
\newblock \href {https://arxiv.org/abs/1908.10084} {Sentence-bert: Sentence
  embeddings using siamese bert-networks}.
\newblock In \emph{Proceedings of the 2019 Conference on Empirical Methods in
  Natural Language Processing}. Association for Computational Linguistics.

\bibitem[{Reimers and Gurevych(2020)}]{reimers2020making}
Nils Reimers and Iryna Gurevych. 2020.
\newblock Making monolingual sentence embeddings multilingual using knowledge
  distillation.
\newblock \emph{arXiv preprint arXiv:2004.09813}.

\bibitem[{Sanh et~al.(2020)Sanh, Debut, Chaumond, and
  Wolf}]{sanh2020distilbert}
Victor Sanh, Lysandre Debut, Julien Chaumond, and Thomas Wolf. 2020.
\newblock \href {http://arxiv.org/abs/1910.01108} {Distilbert, a distilled
  version of bert: smaller, faster, cheaper and lighter}.

\bibitem[{Tahaei et~al.(2021)Tahaei, Charlaix, Nia, Ghodsi, and
  Rezagholizadeh}]{tahaei2021kroneckerbert}
Marzieh~S. Tahaei, Ella Charlaix, Vahid~Partovi Nia, Ali Ghodsi, and Mehdi
  Rezagholizadeh. 2021.
\newblock \href {http://arxiv.org/abs/2109.06243} {Kroneckerbert: Learning
  kronecker decomposition for pre-trained language models via knowledge
  distillation}.

\bibitem[{Vaswani et~al.(2017)Vaswani, Shazeer, Parmar, Uszkoreit, Jones,
  Gomez, Kaiser, and Polosukhin}]{vaswani2017attention}
Ashish Vaswani, Noam Shazeer, Niki Parmar, Jakob Uszkoreit, Llion Jones,
  Aidan~N Gomez, \L~ukasz Kaiser, and Illia Polosukhin. 2017.
\newblock \href
  {https://proceedings.neurips.cc/paper_files/paper/2017/file/3f5ee243547dee91fbd053c1c4a845aa-Paper.pdf}
  {Attention is all you need}.
\newblock In \emph{Advances in Neural Information Processing Systems},
  volume~30. Curran Associates, Inc.

\bibitem[{Wang et~al.(2022)Wang, Qian, Pan, Qi, Kunc, and
  Potdar}]{wang-etal-2022-benchmarking}
Gengyu Wang, Cheng Qian, Lin Pan, Haode Qi, Ladislav Kunc, and Saloni Potdar.
  2022.
\newblock \href {https://doi.org/10.18653/v1/2022.mia-1.7} {Benchmarking
  language-agnostic intent classification for virtual assistant platforms}.
\newblock In \emph{Proceedings of the Workshop on Multilingual Information
  Access (MIA)}, pages 69--76, Seattle, USA. Association for Computational
  Linguistics.

\bibitem[{Wang et~al.(2021)Wang, Zhang, and Han}]{wang2021spatten}
Hanrui Wang, Zhekai Zhang, and Song Han. 2021.
\newblock \href {http://arxiv.org/abs/2012.09852} {Spatten: Efficient sparse
  attention architecture with cascade token and head pruning}.

\bibitem[{Wang et~al.(2020{\natexlab{a}})Wang, Li, Khabsa, Fang, and
  Ma}]{wang2020linformer}
Sinong Wang, Belinda~Z. Li, Madian Khabsa, Han Fang, and Hao Ma.
  2020{\natexlab{a}}.
\newblock \href {http://arxiv.org/abs/2006.04768} {Linformer: Self-attention
  with linear complexity}.

\bibitem[{Wang et~al.(2020{\natexlab{b}})Wang, Wei, Dong, Bao, Yang, and
  Zhou}]{wang2020minilm}
Wenhui Wang, Furu Wei, Li~Dong, Hangbo Bao, Nan Yang, and Ming Zhou.
  2020{\natexlab{b}}.
\newblock \href {http://arxiv.org/abs/2002.10957} {Minilm: Deep self-attention
  distillation for task-agnostic compression of pre-trained transformers}.

\bibitem[{Xiao et~al.(2023)Xiao, Lin, Seznec, Wu, Demouth, and
  Han}]{xiao2023smoothquant}
Guangxuan Xiao, Ji~Lin, Mickael Seznec, Hao Wu, Julien Demouth, and Song Han.
  2023.
\newblock \href {http://arxiv.org/abs/2211.10438} {Smoothquant: Accurate and
  efficient post-training quantization for large language models}.

\bibitem[{Ye et~al.(2021)Ye, Lin, Huang, and
  Sun}]{DBLP:journals/corr/abs-2105-11618}
Deming Ye, Yankai Lin, Yufei Huang, and Maosong Sun. 2021.
\newblock \href {http://arxiv.org/abs/2105.11618} {{TR-BERT:} dynamic token
  reduction for accelerating {BERT} inference}.
\newblock \emph{CoRR}, abs/2105.11618.

\bibitem[{Zhang et~al.(2023)Zhang, Liang, Zhan, Wu, and Lam}]{zhang2023revisit}
Haode Zhang, Haowen Liang, Liming Zhan, Xiao-Ming Wu, and Albert Y.~S. Lam.
  2023.
\newblock \href {http://arxiv.org/abs/2306.05278} {Revisit few-shot intent
  classification with plms: Direct fine-tuning vs. continual pre-training}.

\bibitem[{Zhang et~al.(2022{\natexlab{a}})Zhang, Liang, Zhang, Zhan, Wu, Lu,
  and Lam}]{zhang2022finetuning}
Haode Zhang, Haowen Liang, Yuwei Zhang, Liming Zhan, Xiao-Ming Wu, Xiaolei Lu,
  and Albert Y.~S. Lam. 2022{\natexlab{a}}.
\newblock \href {http://arxiv.org/abs/2205.07208} {Fine-tuning pre-trained
  language models for few-shot intent detection: Supervised pre-training and
  isotropization}.

\bibitem[{Zhang et~al.(2021)Zhang, Zhang, Zhan, Chen, Shi, Wu, and
  Lam}]{zhang2021effectiveness}
Haode Zhang, Yuwei Zhang, Li-Ming Zhan, Jiaxin Chen, Guangyuan Shi, Xiao-Ming
  Wu, and Albert Y.~S. Lam. 2021.
\newblock \href {http://arxiv.org/abs/2109.05782} {Effectiveness of
  pre-training for few-shot intent classification}.

\bibitem[{Zhang et~al.(2022{\natexlab{b}})Zhang, Hashimoto, Wan, Liu, Liu,
  Xiong, and Yu}]{zhang2022pretrained}
Jianguo Zhang, Kazuma Hashimoto, Yao Wan, Zhiwei Liu, Ye~Liu, Caiming Xiong,
  and Philip~S. Yu. 2022{\natexlab{b}}.
\newblock \href {http://arxiv.org/abs/2106.04564} {Are pretrained transformers
  robust in intent classification? a missing ingredient in evaluation of
  out-of-scope intent detection}.

\end{thebibliography}

% \appendix

% \section{Example Appendix}
% \label{sec:appendix}

% This is an appendix.

\end{document}